\pdfoutput=1

\documentclass[11pt]{article}

\usepackage{acl}

\usepackage{times}
\usepackage{latexsym}

\usepackage[T1]{fontenc}

\usepackage[utf8]{inputenc}

\usepackage[T1]{fontenc}    
\usepackage{hyperref}       
\usepackage{url}            
\usepackage{booktabs}       
\usepackage{amsfonts}       
\usepackage{nicefrac}       
\usepackage{microtype}      
\usepackage{xcolor}         
\usepackage{bbding}
\usepackage{graphicx}
\usepackage{makecell}
\usepackage{amsmath}

\usepackage{multirow}
\usepackage{xspace}

\usepackage{threeparttable}

\usepackage[ruled,linesnumbered]{algorithm2e}

\usepackage{amssymb}
\usepackage{amsthm}
\usepackage{ulem}
\theoremstyle{definition}

\usepackage{threeparttable}
\usepackage{caption}
\usepackage{subcaption}

\newcommand{\std}{\scriptsize$\pm$}

\newcommand{\solution}[0]{{Multi-Splits}\xspace}

\makeatletter
\newtheorem*{rep@theorem}{\rep@title}
\newcommand{\newreptheorem}[2]{%
\newenvironment{rep#1}[1]{%
 \def\rep@title{#2 \ref{##1}}%
 \begin{rep@theorem}}%
 {\end{rep@theorem}}}
\makeatother

\newtheorem{theorem}{Finding}
\newreptheorem{theorem}{Finding}

\usepackage{enumitem}
\setenumerate[1]{itemsep=0pt,partopsep=0pt,parsep=\parskip,topsep=2pt,leftmargin=0.4cm}
\setitemize[1]{itemsep=0pt,partopsep=0pt,parsep=\parskip,topsep=2pt,leftmargin=0.4cm}
\setdescription{itemsep=0pt,partopsep=0pt,parsep=\parskip,topsep=2pt,leftmargin=0.4cm}
\usepackage{titlesec}
\titlespacing*{\paragraph} {2pt}{2pt}{5pt}
\usepackage{array}
\usepackage{booktabs}

\newcommand{\midsepdefault}{\aboverulesep = 0.4mm \belowrulesep = 0.4mm} 
\midsepdefault

\title{FewNLU: Benchmarking State-of-the-Art Methods\\ for Few-Shot Natural Language Understanding}

\author{
    Yanan Zheng\thanks{$\;$ The authors have contributed equally to this work.}$\;^{12}$, Jing Zhou\footnotemark[1]$\;^1$, Yujie Qian$^3$, Ming Ding$^1$, Chonghua Liao$^1$\\
    \textbf{Jian Li$^1$, Ruslan Salakhutdinov$^4$, Jie Tang\thanks{$\;$ Corresponding Authors.}$\;^{12}$, Sebastian Ruder\footnotemark[2] $\;^5$, Zhilin Yang\footnotemark[2]$\;^{16}$} \\
    $^1$Tsinghua University, $^2$BAAI, $^3$MIT CSAIL, \\ $^4$Carnegie Mellon University, $^5$Google Research, $^6$Shanghai Qi Zhi Institute \\
    \texttt{\{zyanan, jietang, zhiliny\}@tsinghua.edu.cn}, \\
    \texttt{zhouj18@mails.tsinghua.edu.cn}, \texttt{ruder@google.com} 
}

\begin{document}
\maketitle

\begin{abstract}
The few-shot natural language understanding (NLU) task has attracted much recent attention.
However, prior methods have been evaluated under a disparate set of protocols, which hinders fair comparison and measuring progress of the field.
To address this issue, we introduce an evaluation framework that improves previous evaluation procedures in three key aspects, i.e., test performance, dev-test correlation, and stability.
Under this new evaluation framework, we re-evaluate several state-of-the-art few-shot methods for NLU tasks.
Our framework reveals new insights: 
(1) both the absolute performance and relative gap of the methods were not accurately estimated in prior literature; 
(2) no single method dominates most tasks with consistent performance; 
(3) improvements of some methods diminish with a larger pretrained model;
and (4) gains from different methods are often complementary and the best combined model performs close to a strong fully-supervised baseline.
We open-source our toolkit, FewNLU, that implements our evaluation framework along with a number of state-of-the-art methods.
\footnote{Leaderboard: \url{https://fewnlu.github.io}}
\footnote{Code available at \url{https://github.com/THUDM/FewNLU}}
\end{abstract}

\section{Introduction}

Few-shot learning for natural language understanding (NLU) has been significantly advanced by pretrained language models~\cite[PLMs;][]{gpt3, SchickS21_PET,Schick2021ItsNJ}. 
With the goal of learning a new task with very few (usually less than a hundred) samples, few-shot learning benefits from the prior knowledge stored in PLMs. 
Various few-shot methods based on PLMs and prompting have been proposed~\cite{gptunderstandstoo,adapet,betterfewshotlearner}.

Although the research of few-shot NLU is developing rapidly, the lack of a standard evaluation protocol has become an obstacle hindering fair comparison between various methods on a common ground and measuring progress of the field.
While some works~\cite{Schick2021ItsNJ,adapet}
experimented with a fixed set of hyper-parameters, prior work \cite{truefew-shot,revisitingfewsamplefinetune} noted that such a setting might be exposed to the risk of overestimation~.\footnote{This is because the fixed hyper-parameters are selected according to practical considerations, which are informed by the test set performance from previous evaluations.}
Other works~\cite{gptunderstandstoo,betterfewshotlearner,truefew-shot} proposed to use a small development set to select hyper-parameters,
but their evaluation protocols vary in a few key aspects (e.g., how to construct data splits), which in fact lead to large differences as we will show (in Section \ref{sec:howtosplit}).
The above phenomena highlight the need for a common protocol for the evaluation of few-shot NLU methods.
However, the fact that few-shot learning is extremely sensitive to subtle variations of many factors~\cite{weightinitialization,betterfewshotlearner} poses challenges for designing a solid evaluation protocol.

In this work, aiming at addressing the aforementioned challenge, we propose an evaluation framework for few-shot NLU. The evaluation framework consists of a repeated procedure---selecting a hyper-parameter, selecting a data split, training and evaluating the model. To set up a solid evaluation framework, 
it is crucial to specify a key design choice---how to construct data splits for model selection. We conduct a comprehensive set of experiments to answer the question.
Specifically, we propose a ``\solution'' strategy, which randomly splits the available labeled samples into training and development sets multiple times, followed by aggregating the results from each data split. 
We show that this simple strategy outperforms several baseline strategies 
in three dimensions: (1) the test set performance of the selected hyper-parameters; (2) correlation between development set and true test set performance; and (3) robustness to hyper-parameter settings.


We then take a step further to re-evaluate recent state-of-the-art few-shot NLU methods under this common evaluation framework. Our re-evaluation leads to several key findings summarized in Section~\ref{sec:findings}.

To aid reproducing our results and benchmarking few-shot NLU methods, we open-source FewNLU, a toolkit that contains implementations of a number of state-of-the-art methods, data processing utilities, as well as our proposed evaluation framework.

To sum up, our contributions are as follows.
\begin{enumerate}
    \item We introduce a new evaluation framework of few-shot NLU. We propose three desiderata of few-shot evaluation and show that our framework outperforms previous ones in these aspects.
    Thus our framework allows for more reliable comparison of few-shot NLU methods.

    \item Under the new evaluation framework, we benchmark the performance of recent methods individually as well as the best performance with a combined approach. These benchmarks reflect the current state of the art and will serve as important baselines for future research.
    
    \item Throughout our exploration, we arrive at several key findings summarized in Section \ref{sec:findings}.

    \item We open-source a toolkit, FewNLU, to facilitate future research with our framework. 

\end{enumerate}

\section{Summary of Findings} \label{sec:findings}

For reference, we collect our key findings here and discuss each of them throughout the paper.

\begin{reptheorem}{thm:multisplits}
Our proposed \solution is a more reliable data-split strategy than several baselines with improvements in (1) test performance, (2) correlation between development and test sets, and (3) stability w.r.t. the number of runs.
\end{reptheorem}


\begin{reptheorem}{thm:decrease}
The absolute performance and the relative gap of few-shot methods were in general not accurately estimated in prior literature.
It highlights the importance of evaluation for obtaining reliable conclusions.
Moreover, the benefits of some few-shot methods (e.g., ADAPET) decrease on larger pretrained models.
\end{reptheorem}

\begin{reptheorem}{thm:combinationsota}
Gains of different methods are largely complementary. A combination of methods largely outperforms individual ones, performing close to a strong fully-supervised baseline with RoBERTa.
\end{reptheorem}

\begin{reptheorem}{thm:combination}
No single few-shot method dominates most NLU tasks. This highlights the need for the development of few-shot methods with more consistent and robust performance across tasks.
\end{reptheorem}

\section{Related Work}\label{sec:relatedwork}

The pretraining-finetuning paradigm~\cite{Howard2018ulmfit} shows tremendous success in few-shot NLU tasks.
Various methods have been developed such as [CLS] classification~\cite{bertpaper}, prompting-based methods with discrete prompts~\cite{Schick2021ItsNJ,betterfewshotlearner} or continuous prompts~\cite{gptunderstandstoo, AutoPrompt, prefix-tuning, Promptuning}, and methods that calibrate the output distribution~\cite{freelunchforfewshot,calibratebeforeuse}.

The fact that few-shot learning is sensitive to many factors and thus is extremely unstable~\cite{whatmakesgoodincontext, fantasticallyorderdprompts, revisitingfewsamplefinetune,weightinitialization} increases the difficulty of few-shot evaluation.
Several works address evaluation protocols to mitigate the effects of instability: 
~\citet{betterfewshotlearner} and \citet{gptunderstandstoo} adopt a held-out set to select models.
~\citet{truefew-shot} proposed $K$-fold cross-validation and minimum description length evaluation strategies. 
Our work differs from these works on few-shot evaluation in several aspects:
(1) we propose three metrics to evaluate data split strategies; (2) while most prior work proposed evaluation protocols without justification, we conduct comprehensive experiments to support our key design choice;
(3) we formulate a general evaluation framework;
(4) our re-evaluation under the proposed framework leads to several key findings.



Though there have been a few existing few-shot NLP benchmarks, our work is quite different in terms of the key issues addressed.
FLEX~\cite{flex} and CrossFit~\cite{crossfit} studied principles of designing tasks, datasets, and metrics.
FewGLUE~\cite{Schick2021ItsNJ} is a dataset proposed for benchmarking few-shot NLU.
CLUES~\cite{CLUES} pays attention to the unified format, metric, and the gap between human and machine performance.
While the aforementioned benchmarks focus on ``what data to use'' and ``how to define the task'', our work discussed ``how to evaluate'' which aims at establishing a proper evaluation protocol for few-shot NLU methods.
Since FewNLU is orthogonal to the aforementioned prior work, it can also be employed on the data and tasks proposed in previous work.

\section{Evaluation Framework}

Formally, for a few-shot NLU task, we have a small labeled set ${\mathcal{D}}_{\text{label}}=\{(x_i, y_i)\}_{i}^N$ and a large test set ${\mathcal{D}}_{\text{test}}=\{(x_i^{\text{test}}, y_i^{\text{test}})\}_{i}$ where $N$ is the number of labeled samples, $x_i$ is a text input (consisting of one or multiple pieces), and $y_i \in \mathcal{Y}$ is a label.
The goal is to finetune a pretrained model with $\mathcal{D}_{\text{label}}$ to obtain the best performance on $\mathcal{D}_{\text{test}}$. An unlabeled set $\mathcal{D}_{\text{unlab}}=\{x_i^{\text{unlab}}\}_i$ may additionally be used by semi-supervised few-shot methods (\textsection \ref{sec:few-shot-methods}).

\subsection{Formulation of Evaluation Framework}\label{sec:formulation}

Our preliminary results (in Appendix \textsection\ref{sec:doweneedtoselect}) show that using a fixed set of hyper-parameters~\cite{SchickS21_PET,Schick2021ItsNJ} is sub-optimal, and model selection is required.
It motivates us to study a more robust evaluation framework for few-shot NLU.
The goal of an evaluation framework is twofold: (1) benchmarking few-shot methods for NLU tasks such that they can be fairly compared and evaluated; and (2) obtaining the best few-shot performance in practice.
In light of the two aspects, we propose the few-shot evaluation framework in Algorithm~\ref{alg:framework}.

The framework searches over a hyper-parameter space $\mathcal{H}$ to evaluate a given few-shot method $M$, obtaining the best hyper-parameter setting $h^\star$ and its test set results.
\footnote{For simplicity and ease of use, we use grid search for searching the hyper-parameter space $\mathcal{H}$ and identify critical hyper-parameters to limit its size. More complex search methods such as Bayesian Optimization \cite{snoek2012practical} could be used to search over larger hyper-parameter spaces.}
The measurement for each $h$ is estimated by performing training and evaluation on multiple data splits (obtained by splitting $\mathcal{D}_{\text{label}}$ according to a strategy) and reporting their average dev set results.
Finally, the method is evaluated on $\mathcal{D}_{\text{test}}$ using the checkpoints corresponding to $h^\star$.
For benchmarking, we report the average and standard deviation over multiple test set results. 
Otherwise, that is, to achieve a model with the best practical performance, we re-run on the entire $\mathcal{D}_{\text{label}}$ with $h^\star$.

\begin{algorithm}
\caption{\small{A Few-Shot Evaluation Framework}}
\label{alg:framework}
\small
\KwData{
$\mathcal{D}_{\text{label}}$, $\mathcal{D}_{\text{test}}$,
a hyper-parameter space $\mathcal{H}$,
a few-shot method $M$,
the number of runs $K$.}
\KwResult{test performance; best hyper-parameter $h^\star$.}
\BlankLine
\For{$k\leftarrow 1\cdots K$}{
{Divide $\mathcal{D}_{\text{label}}$ into $\mathcal{D}_{\text{train}}^{k}$ and $\mathcal{D}_{\text{dev}}^{k}$ according to a data-split strategy}\;
}
  \For{$h \in \mathcal{H}$}{
    \For{$k\leftarrow 1 \cdots K$}{
    {Run the method $M$ by training on $\mathcal{D}_{\text{train}}^{k}$ and evaluating on $\mathcal{D}_{\text{dev}}^{k}$}\;
    {Report the dev-set performance $\mathcal{P}_{\text{dev}}^{h,k}$.}
    }
    {Compute the mean and standard deviation over $K$ dev-set results, $\mathcal{P}_{\text{dev}}^h \pm \mathcal{S}_{\text{dev}}^h$}\;
  }
  Select $h^\star$ with the best $\mathcal{P}_{\text{dev}}^h$.\;
  \uIf{the goal is to evaluate a method}
  {
    Evaluate on the test set $\mathcal{D}_{\text{test}}$ with the $K$ checkpoints that correspond to $h^\star$\;
    Report the mean and standard deviation over the $K$ test results $\mathcal{P}_{\text{test}}^{h\star} \pm \mathcal{S}_{\text{test}}^{h\star}$.
  }
  \uElseIf{the goal is to obtain the best performance}{
  Re-run on the entire $\mathcal{D}_{\text{label}}$ using fixed $h^\star$ with $L$ different random seeds\;
  Evaluate on the test set with the $L$ checkpoints\;
  Report the mean and stddev over $L$ test results.
  }
  \textbf{end}
\end{algorithm}

The framework requires specifying a key design choice---how to construct the data splits, which we will discuss in \textsection\ref{sec:howtosplit}.


\subsection{How to Construct Data Splits}\label{sec:howtosplit}

\subsubsection{Desiderata: Performance, Correlation, and Stability}

We first propose the following three key desiderata for the evaluation of different data split strategies.
\begin{enumerate}
    \item 
    \textbf{Performance of selected hyper-parameter.}
    A good data split strategy should select a hyper-parameter that can achieve a good test set performance.
    We use the same metrics as~\cite{Schick2021ItsNJ}, along with standard deviations.

    \item \textbf{Correlation between dev and test sets (over a hyper-parameter distribution)}.
    Since a small dev set is used for model selection, it is important for a good strategy to obtain a high correlation between the performances on the small dev set and test set over a distribution of hyper-parameters.
    We report the Spearman's rank correlation coefficient for measurement. 

    \item \textbf{Stability w.r.t. number of runs $K$.}
    The choice of the hyper-parameter $K$ should have small impacts on the above two metrics (i.e., performance and correlation).
    To analyze the stability w.r.t $K$, we report the standard deviation over multiple different values of $K$.
    Besides, it is desirable to have reduced variance when $K$ increases.
    Thus we report the above two metrics with different values of $K$ and the standard deviation of test scores over $K$ runs.

\end{enumerate}

\subsubsection{Data Split Strategies}\label{sec:datasplits}

We consider several data split strategies.
Some are proposed by previous work, including $K$-fold cross validation~(CV)~\cite{truefew-shot},
minimum description length (MDL)~\cite{truefew-shot}, and bagging (BAG)~\cite{bagging}.
We also consider two simple strategies worth exploring, including random sampling (RAND) and model-informed splitting (MI).
And we propose a new data split strategy, \solution (MS).
Besides, we also experiment a special case of CV when $K$ equals the number of labeled sample, which is leave-of-out cross validation (LOOCV).
Since LOOCV takes much longer time and suffers from efficiency problem, we only experimented on several tasks and left the results in Appendix~\ref{apdx:loocv}.
They all fit into the pipeline of the proposed framework in \textsection\ref{sec:formulation}:
\begin{enumerate}
    \item \textbf{$K$-fold CV} 
    equally partitions $\mathcal{D}_{\text{label}}$ into $K$ folds.
    Each time, it uses the $k^{\text{th}}$ fold for validation and the other $K-1$ folds for training.

    \item \textbf{MDL} assigns half of $\mathcal{D}_{\text{label}}$ as the joint training data and equally partitions the other half into $K$ folds.
    Each time, it uses the $k^{\text{th}}$ fold for validation, and all its previous $k-1$ folds together with the joint training data for training. 
    
    \item \textbf{Bagging} samples $N\times r$ ($r \in (0, 1]$ is a fixed ratio) examples with replacement from the labeled sample as the training set, leaving samples that do not appear in the train set for validation.
    
    \item \textbf{Random Sampling} performs random sampling without replacement from $\mathcal{D}_{\text{label}}$ twice, respectively sampling $N\times r$ and $N\times (1-r)$ data as the training and development sets.
    
    \item \textbf{Model-Informed Splitting} computes representations of each labeled example using a model, and clusters them into two distinct sets, respectively as the training and development sets. \footnote{Specifically, we used a BERT-Base model to encode data and take the [CLS] representations.}
    
    \item \textbf{\solution} randomly splits $\mathcal{D}_{\text{label}}$ into training and development sets using a fixed split ratio $r$.
    
\end{enumerate}

\begin{table}[]
    \centering
    \tiny
    \setlength{\tabcolsep}{1.5mm}
    \resizebox{\linewidth}{!}{%
    \begin{tabular}{l|l|l}
    \toprule[1pt]
    & \#Train & \#Dev \\
    \midrule
    CV              & $(K-1) \times N / K$ & $N / K$ \\
    MDL             & $N/2+N(k-1)/(2K)$ & $N/(2K$) \\
    BAG         & $N \times r$  & $\textgreater (N \times (1-r))$ \\
    RAND            & $N \times r$  & $N \times (1-r)$ \\
    Multi-Splits &     $N\times r$     & $N \times (1-r)$ \\
    \bottomrule[1pt]
    \end{tabular}}
    \caption{{Number of examples of training and development sets for different strategies. $N$: number of labeled data, $K$: number of runs, $k$: the $k^{\text{th}}$ split for MDL; $r$: split ratio. MI is omitted since its number of examples depends on the dataset.}}
    \label{tab:numberofdata}
\end{table}

\begin{table*}[!h]
\centering
\resizebox{\textwidth}{!}{%
  \begin{tabular}{l|cccccccccc}
    \toprule[1pt]  
    \multirow{2}*{}
    & BoolQ & RTE & WiC & \multicolumn{2}{c}{CB} & \multicolumn{2}{c}{MultiRC} & WSC & COPA & Avg. \\
    & Acc. & Acc. & Acc. & Acc. & F1 & F1a & EM. &  Acc. & Acc \\
    \midrule
    CV 
    & 82.71 \std1.29
    & 77.80 \std2.25
    & 64.42 \std1.63
    & 90.18 \std2.31
    & 87.52 \std2.20
    & 80.08 \std1.15
    & 45.02 \std1.46
    & 82.45	\std3.71
    & 92.25	\std1.71
    & 78.72 \\
    MDL 
    & 76.43 \std7.12 
    & 83.94 \std1.49
    & 63.68 \std3.38
    & 84.38 \std5.13
    & 82.03 \std5.69
    & 77.63 \std1.20
    & 43.81 \std1.32
    & 81.49 \std3.95
    & 89.50 \std3.32
    & 77.00
    \\
    BAG 
    & 81.77 \std1.48
    & 77.98 \std1.56
    & 65.56 \std3.26
    & 87.50 \std6.90
    & 77.15 \std13.76
    & 79.62 \std1.26
    & 43.60 \std1.98
    & 85.34 \std2.87
    & 88.75 \std3.10
    & 77.62
    \\
    RAND 
    & 78.79	\std5.40
    & 82.13	\std0.91
    & 59.60 \std3.89
    & 86.16 \std3.05
    & 74.04	\std12.94
    & 80.14 \std2.20
    & 44.88 \std4.45
    & 84.38	\std2.99
    & 90.75	\std3.59
    & 76.89 \\
    MI 
    & 78.25	\std1.59
    & 77.35	\std4.06
    & 64.66 \std1.48
    & 88.84 \std1.71
    & 84.75	\std4.32
    & 76.75 \std0.44
    & 40.95 \std0.10
    & 83.41	\std6.00
    & 78.75	\std8.06
    & 75.44 \\
    \midrule[1pt]
    MS
    & 82.67	\std0.78
    & 79.42	\std2.41
    & 67.20 \std1.34
    & 91.96 \std3.72			
    & 88.63	\std4.91
    & 78.20 \std1.86
    & 42.42 \std3.04
    & 84.13 \std4.87
    & 89.00 \std2.94
    & 79.00 \\
    \bottomrule[1pt]
\end{tabular}
}
\caption{Test performance of different data-split strategies with PET on FewGLUE ($K$=4).
Larger scores means the strategy effectively selects a model that achieves better test set performance.}
\label{tab:fewshotperformance}
\end{table*}

\begin{table*}
\centering
\resizebox{0.67\linewidth}{!}{%
  \begin{tabular}{l|cccccccc}
    \toprule[1pt]  
     & BoolQ & RTE & WiC & CB & MultiRC & WSC & COPA & Avg. \\
    \midrule  
    CV
    & -0.0497
    & 0.8561
    & 0.8184	
    & 0.5286	
    & 0.2283	
    & 0.1507	
    & 0.5668	
    & 0.4427
    \\
    MDL
    & -0.1143
    & 0.7806	
    & 0.6326
    & 0.3274	
    & 0.1910
    & 0.1278
    & 0.6342	
    & 0.3685
    \\
    BAG
    & 0.5533	
    & 0.8714	
    & 0.9572	
    & 0.6809	
    & 0.6340	
    & 0.2550	
    & 0.7491	
    & 0.6716
    \\
    RAND
    & 0.7453
    & 0.7602
    & 0.8048	
    & 0.6764	
    & 0.3253	
    & 0.0795	
    & 0.9004	
    & 0.6131
    \\
    MI
    & 0.5651	
    & 0.6832
    & 0.7780	
    & 0.6618	
    & 0.6651	
    & 0.0200	
    & 0.5902
    & 0.5662\\
    \midrule[1pt]
    MS 
    & 0.7079
    & 0.8266
    & 0.9464
    & 0.7558
    & 0.4983	
    & 0.3986
    & 0.8997	
    & 0.7190
    \\
    \bottomrule[1pt]
\end{tabular}
}
\caption{{Correlation results of different data-split strategies with PET on FewGLUE ($K$=4). Larger values means the strategy is better at selecting the best test results using dev sets.}}
\label{tab:fewshotcorrelation}
\end{table*}



Essentially, these data split strategies differ in several key aspects.
\begin{enumerate}
\item For CV and MDL, $K$ controls the number of runs and the split ratio.
For \solution, BAG and RAND, the split ratio is decoupled from $K$ and is controlled by $r$.
For MI, the split ratio and number of runs depend on $\mathcal{D}_{\text{label}}$.

\item They use a different amount of data for training and development sets as Table~\ref{tab:numberofdata} shows.

\item There are cases when CV and MS share the same split ratio. The difference is that MS allows overlap between splits while CV does not.

\item BAG allows duplicated training data, while RAND and \solution do not.
The training and development sets do not overlap for BAG and \solution but overlap for RAND.

\end{enumerate}

In the limit, our \solution is similar to leave-$P$-out cross-validation \cite[LPOCV;][]{celisse2014optimal}\footnote{Leave-$P$-out cross-validation uses $P$ data examples as the development set and the remaining data examples as the training set. This is repeated on all ways to cut the labeled dataset in a development set and a training set.
} where LPOCV runs $\binom{N}{P}$ times ($P$ is the number of dev set examples) while \solution runs $K$ times.
As $K$ increases, \solution gradually approaches LPOCV.
Since it is impossible to enumerate the large number of possible splits in practice, \solution can be viewed as a practical version of LPOCV.
Compared to the strategy of~\cite{betterfewshotlearner} that uses multiple datasets, our \solution uses multiple data splits for a single dataset. It is thus more practical as in real-world scenarios, it is hard to obtain multiple labeled datasets for a true few-shot problem; 
otherwise, it could be formulated as a fully-supervised learning problem.
The strategy in~\cite{gptunderstandstoo} is a special case of \solution when $K=1$, which suffers from higher variance.

\subsubsection{Experimental Setup}\label{sec:setup}

To evaluate different data split strategies, we experiment on the FewGLUE benchmark~\cite{Schick2021ItsNJ}.
We evaluate strategies based on the widely used prompt-based few-shot method PET~\cite{Schick2021ItsNJ} with DeBERTa as the base model.\footnote{We fixed the parameters of DeBERTa's bottom one-third layers due to GPU memory limitations, which did not affect the performance much in our preliminary experiments.}
We run experiments on the same tasks with the same hyper-parameter space to ensure a fair comparison; in this experiment we search learning rate, evaluation ratio, prompt pattern and maximum training step. More experimental details are in Appendix~\ref{apdx:part1_details}.

\subsubsection{Main Results and Analysis}\label{sec:settingmainresults}

\begin{figure}
    \centering
    {\includegraphics[width=\linewidth] {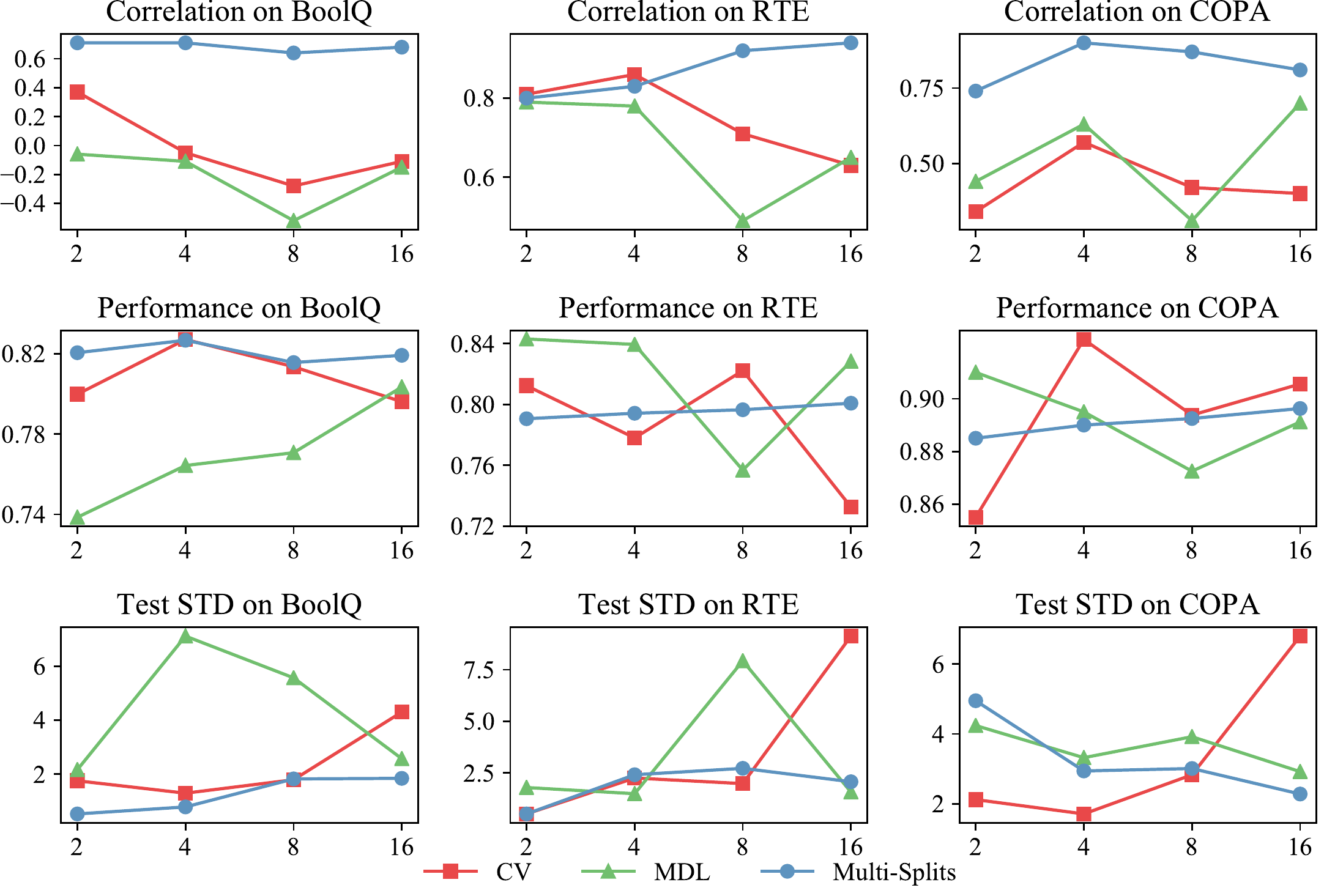}}
    \caption{{Test performance, correlation and standard deviation along with different $K$ on BoolQ, RTE, and COPA tasks under different strategies. A smooth and stable dot-line indicates the setting is insensitive to the choice of $K$.}} 
    \label{fig:Kresults}
\end{figure}

Table~\ref{tab:fewshotperformance}, Table~\ref{tab:fewshotcorrelation} and Figure~\ref{fig:Kresults} show the main results with 64 labeled samples. 

It is noteworthy that we also experimented with 32 labeled samples and have observed that varying the number of labeled examples does not affect the following conclusion (see Appendix~\ref{apdx:part1_details}).

\paragraph{Test Performance and Correlation.}
From both Table~\ref{tab:fewshotperformance} and Table~\ref{tab:fewshotcorrelation}, we find that \solution achieves the best average test set performance as well as the best average correlation among all strategies. 
We analyze them as follows:\footnote{In the following explanation, the numbers refer to the total training/development data covering $K$=4 runs.}

\begin{enumerate}
\item \solution uses fewer labeled samples for training (i.e., 128) while CV and MDL use more (i.e., 192 and 176).
Despite using more training data, both CV and MDL do not perform better. This indicates few-shot performance is limited by not being able to select the best model rather than not having sufficient training data. 
Both CV and MDL use fewer data for validation (i.e., 64 and 32) than \solution (i.e., 128), thus leading to poor correlation. 

\item Although \solution and BAG use the same number of training data (i.e., 128), there could be duplication in the training set of BAG, making it poor in diversity and further leading to lower test performance, compared to \solution. This indicates diversity of training sets is crucial when constructing few-shot data splits.

\item RAND uses similar-sized dev and train sets to BAG and MS but performs worse in test performance.
Since there could be overlap between train and dev sets, the model may have memorized data, leading to poor test performance.

\item MI constructs very different train and dev sets. Overfitting on one of them and validating on the other pose more challenges for the few-shot method on out-of-distribution tasks.
\end{enumerate}

\paragraph{Stability w.r.t. the number of runs $K$.}
Figure~\ref{fig:Kresults} shows the results on stability.
In light of limited computation resources, we only experiment with some representative strategies.
Both CV and MDL represent strategies whose number of runs are coupled with the size of data split, while \solution represents strategies that have a fixed ratio and independent $K$.
We observe:
(1) \solution (blue lines) is the most stable in correlation and performance, while other strategies CV and MDL 
are more sensitive to the choice of $K$.
(2) \solution shows the smallest variance over multiple runs on both BoolQ and RTE.
For COPA, though \solution shows high variance when $K=2$, the variance becomes smaller with larger $K$, while CV and MDL suffer from increasing or unstable variance.

A possible explanation is that increasing $K$ does not affect the number of training and development examples for \solution; instead, it increases the confidence of results. An important practical benefit of \solution is that one can always choose to increase $K$ for lower variance.
However, for CV and MDL, the sizes of training and development sets are affected by $K$, where extremely large $K$ value leads to a failure mode and extremely small $K$ leads to unstable results.
In practice, it is hard to know which value of $K$ to use a priori.

To sum up, based on the aforementioned results and analysis, 
we arrive at the following finding.
\begin{theorem}\label{thm:multisplits}
Our proposed \solution is a more reliable data-split strategy than several baselines with improvements in (1) test performance, (2) correlation between development and test sets, and (3) stability w.r.t. number of runs.
\end{theorem}

\paragraph{Remark}
Our evaluation framework is better in terms of test performance, dev-test correlation, and stability, which proves it can achieve possible peak performance, reliably select the corresponding hyperparameters according to dev results without overfitting, and mitigate the effects of randomness to the maximum extent. Therefore, the estimation of our evaluation framework for model performance is more reliable than previous evaluations.
\section{Re-Evaluation of State-of-the-Art Methods}\label{sec:reevaluation}

\begin{table*}
\small
\centering
\begin{threeparttable}
\begin{tabular}{p{1cm}<{\centering}|p{1.85cm}|ccccccccccc}
    \toprule[1pt]  
    \multirow{2}*{\shortstack{Base\\ Models}} &
    \multirow{2}*{\shortstack{Few-Shot\\ Methods}}  & BoolQ & RTE & WiC & \multicolumn{2}{c}{CB} & \multicolumn{2}{c}{MultiRC} & WSC & COPA & Avg. \\
    & & Acc. & Acc. & Acc. & Acc. & F1 & F1a & EM. & Acc. & Acc \\
    \midrule
    \multirow{18}*{{\scriptsize ALBERT}}
    & CLS 
    & 55.01 
    & 53.97 
    & 50.82 
    & 67.97 
    & 52.18 
    & 59.95 
    & 18.86 
    & 52.64  
    & 64.25 
    & 53.74\\
    & 
    & \std2.95 
    & \std5.49 
    & \std3.02 
    & \std18.29 
    & \std10.30 
    & \std10.69 
    & \std9.80 
    & \std10.25 
    & \std9.36 
    & \\
    & PET 
    & 76.70 
    & 72.83 
    & 53.87 
    & 84.38 
    & 62.56 
    & 76.51 
    & 36.46 
    & 80.05 
    & 81.75 
    & 70.74 \\
    &
    & \std1.85 
    & \std1.30 
    & \std4.47 
    & \std4.47 
    & \std7.66 
    & \std1.52 
    & \std2.13 
    & \std2.53 
    & \std4.03 
    & \\
    & ADAPET 
    & 79.24 
    & 74.28 
    & 58.07 
    & 92.86 
    & 89.99 
    & 77.24 
    & 37.17 
    & 78.85 
    & 81.75 
    & 74.40 \\
    &
    & \std1.42
    & \std3.57
    & \std2.96
    & \std1.46
    & \std3.91
    & \std1.99
    & \std2.64
    & \std4.51 
    & \std3.95
    & \\
    & P-tuning 
    & 76.55 
    & 63.27 
    & 55.49 
    & 88.39 
    & 84.24 
    & 75.91 
    & 38.01 
    & 78.85 
    & 85.25 
    & 71.81\\
    & 
    & \std2.68
    & \std3.63
    & \std1.21
    & \std3.72
    & \std5.15
    & \std1.74
    & \std0.78
    & \std1.76 
    & \std3.30
    & \\
    & FlipDA 
    & 77.95 
    & 70.85 
    & 57.17 
    & 83.93 
    & 74.30 
    & 76.05 
    & 35.68 
    & 79.57 
    & 87.50 
    & 72.57 \\
    & 
    & \std2.60 
    & \std2.71 
    & \std2.59 
    & \std4.37 
    & \std13.23 
    & \std1.33 
    & \std1.44 
    & \std1.82 
    & \std3.70 \\
    \cline{2-12} & \\[-1.5ex]
    & PET+MLM$^{3}$ 
    & 76.83 
    & 71.48 
    & 52.39 
    & 83.93 
    & 67.37 
    & 75.15 
    & 35.68 
    & 81.97 
    & 85.75 
    & 71.36 \\
    & 
    & \std1.18
    & \std1.64
    & \std1.44
    & \std5.05
    & \std8.31
    & \std0.34
    & \std1.10
    & \std1.82
    & \std3.40 & \\
    & iPET(single)$^{3,4}$ 
    & 74.29 
    & 72.35 
    & 54.78 
    & 84.67 
    & 76.92 
    & 76.33 
    & 37.72 
    & 77.80 
    & 84.00 
    & 71.58 \\
    & 
    & \std4.10
    & \std3.71
    & \std3.93
    & \std3.18
    & \std5.44
    & \std1.18
    & \std2.58
    & \std2.79
    & \std6.02 & \\
    & Noisy(single)$^{3,4}$ 
    & 76.11 
    & 72.62 
    & 54.11 
    & 84.38 
    & 72.57 
    & 76.59 
    & 37.00 
    & 79.17 
    & 83.50 
    & 71.54 \\
    & 
    & \std2.16
    & \std2.80
    & \std1.98
    & \std5.60
    & \std11.84
    & \std1.40
    & \std2.34
    & \std3.31
    & \std3.34 \\
    & iPET(cross)$^{3,4}$ 
    & 76.83 
    & 74.28 
    & 58.35 
    & 83.48 
    & 73.86 
    & 75.71 
    & 37.30 
    & 76.44 
    & 83.25 
    & 72.05 \\
    &
    & \std1.39
    & \std4.31
    & \std2.42
    & \std2.68
    & \std2.48
    & \std2.14
    & \std2.71
    & \std2.78
    & \std4.19 \\
    & Noisy(cross)$^{3,4}$ 
    & 75.64 
    & 75.27 
    & 56.43 
    & 84.82 
    & 77.79 
    & 77.11 
    & 38.25 
    & 80.53 
    & 83.00 
    & 72.84 \\
    & 
    & \std1.82
    & \std1.97
    & \std2.67
    & \std4.49
    & \std8.46
    & \std1.49
    & \std0.92
    & \std7.17
    & \std4.76 \\  
    \midrule[1pt]
    \multirow{20}*{\scriptsize DeBERTa}
    & CLS 
    & 59.49 
    & 49.55 
    & 54.08 
    & 68.30 
    & 60.10 
    & 75.42 
    & 34.23 
    & 53.13
    & 85.25 
    & 60.07 \\
    & 
    & \std1.74 
    & \std2.23 
    & \std2.15 
    & \std3.96 
    & \std10.14 
    & \std2.39
    & \std5.02 
    & \std5.17 
    & \std2.22 & \\
    & PET  
    & 82.67
    & 79.42 
    & \underline{67.20} 
    & 91.96 
    & 88.63 
    & 78.20 
    & 42.42
    & 84.13  
    & 89.00 
    & 79.00\\
    & 
    & \std0.78 
    & \std2.41 
    & \std1.34 
    & \std3.72 
    & \std4.91 
    & \std1.86 
    & \std3.04 
    & \std4.87 
    & \std2.94 \\
    & ADAPET 
    & 81.28 
    & \underline{82.58} 
    & 66.50  
    & 89.73
    &86.63 
    & 77.88 
    & 43.05 
    & \underline{85.34} 
    & 88.75 
    & 79.01 \\
    & 
    & \std1.26 
    & \std2.44 
    & \std2.11 
    & \std6.08 
    & \std7.29 
    & \std2.55 
    & \std3.60 
    & \std2.13 
    & \std4.43 \\
    & P-tuning 
    & 82.25 
    & 82.22 
    & 66.22 
    & 94.20 
    & 91.76  
    & 78.45 
    & 43.78 
    & 85.10 
    & 86.50 
    & 79.48 \\
    & 
    & \std0.85 
    & \std1.23
    & \std1.18 
    & \std2.25
    & \std3.30
    & \std1.46 
    & \std3.93 
    & \std4.87 
    & \std3.70 &\\
    & FlipDA 
    & \underline{83.52} 
    & 80.14 
    & 65.28 
    & \underline{95.09} 
    & \underline{93.57} 
    & \underline{80.21} 
    & \underline{46.67} 
    & \underline{85.34} 
    & \underline{90.50} 
    & \underline{80.37}\\
    &
    & \std 0.35 
    & \std 1.93 
    & \std 1.56 
    & \std 2.68 
    & \std 2.62 
    & \std 1.35 
    & \std 0.82 
    & \std 3.27 
    & \std 1.00 \\
    \cline{2-12}& \\[-1.5ex]
    & PET+MLM$^{3}$ 
    & 82.80 
    & \uwave{83.30} 
    & 58.23 
    & 90.18 
    & 87.18 
    & 77.05 
    & 40.63 
    & 81.73 
    & 85.75 
    & 77.05 \\
    & 
    & \std0.97 
    & \std2.40 
    & \std4.98 
    & \std3.09 
    & \std6.17 
    & \std1.80 
    & \std1.64 
    & \std5.77 
    & \std3.40 \\
    & iPET(single)$^{3,4}$ 
    & 81.27 
    & 81.11 
    & 64.75 
    & 89.88 
    & 87.70 
    & \uwave{79.99} 
    & \uwave{45.23} 
    & 82.93 
    & 90.83 
    & 78.90 \\
    & 
    & \std1.61 
    & \std1.89 
    & \std4.27 
    & \std5.01 
    & \std6.52 
    & \std1.94 
    & \std2.19 
    & \std 3.76 
    & \std2.79 \\
    & Noisy(single)$^{3,4}$ 
    & 81.60 
    & 81.95 
    & 65.97 
    & \uwave{91.67} 
    & 89.17 
    & 79.85 
    & 45.10 
    & 84.46 
    & 90.67 
    & 79.65  \\
    & 
    & \std1.54 
    & \std2.01 
    & \std2.44 
    & \std2.33 
    & \std2.95 
    & \std1.22
    & \std2.58 
    & \std2.49 
    & \std2.53 \\
    & iPET(cross)$^{3,4}$ 
    & \uwave{83.45} 
    & 83.12 
    & \uwave{69.63} 
    & 91.52 
    & \uwave{90.72} 
    & 79.92 
    & 44.96 
    & \uwave{86.30} 
    & \uwave{93.75} 
    & 81.40 \\
    & 
    & \std0.90 
    & \std1.04 
    & \std2.15 
    & \std3.05 
    & \std2.68 
    & \std1.11 
    & \std3.13 
    & \std1.64 
    &\std2.99\\
    & Noisy(cross)$^{3,4}$ 
    & 82.19 
    & 81.95 
    & 68.26 
    & 90.18 
    & 86.74 
    & 79.48 
    & 44.20 
    & 83.41 
    & \uwave{93.75}
    & 79.98\\
    &
    & \std0.65 
    & \std0.51 
    & \std 1.12 
    & \std 2.31 
    & \std 3.00 
    & \std 2.53
    & \std 4.14 
    & \std 4.18 
    & \std 3.30 \\
    \midrule[1pt]
    \multirow{2}*{\scriptsize DeBERTa}
    & Our Best$^{3,4}$ 
    & \textbf{84.0} 
    & \textbf{85.7} 
    & \textbf{69.6} 
    & \textbf{95.1} 
    & \textbf{93.6} 
    & \textbf{81.5} 
    & \textbf{48.0} 
    & \textbf{88.4} 
    & \textbf{93.8} 
    & \textbf{85.44}$^1$ \\
    & \quad (few-shot) 
    & \std0.55 
    & \std0.63 
    & \std2.15 
    & \std2.68  
    & \std2.62 
    & \std0.76
    & \std0.99 
    & \std2.82  
    & \std2.99 &  \\
    {\scriptsize RoBERTa} 
    & \makecell[l]{RoBERTa $^5$ \\\quad (fully sup.)} 
    & 86.9 
    & 86.6 
    & 75.6 
    & 98.2 
    & - 
    & 85.7 
    & - 
    & 91.3 
    & 94.0 
    & 88.33 \\
    {\scriptsize DeBERTa} & \makecell[l]{DeBERTa $^2$ \\\quad (fully sup.)} 
    & 88.3 
    & 93.5 
    & - 
    & - 
    & - 
    & 87.8 
    & 63.6 
    & - 
    & 97.0 
    & - \\
    \bottomrule[1pt]
\end{tabular}
\begin{tablenotes}
     \item[1] For comparison with RoBERTa (fully sup.), the average of Our Best (few-shot) 85.17 excludes MultiRC-EM and CB-F1.
     \item[2] The fully-supervised results on DeBERTa are reported in \url{https://github.com/THUDM/GLM}.
     \item[3] Unlabeled data are used.
     \item[4] The ensemble technique is used.
     \item[5] The RoBERTa (fully-sup.) results by~\cite{liu2019roberta}. RoBERTa-large has less parameters than DeBERTa-xxlarge-v2. 
\end{tablenotes}
\end{threeparttable}
\caption{\small{
Re-evaluation of few-shot methods on ALBERT and DeBERTa under our evaluation framework with \solution strategy on test set of our setup.
For iPET and Noisy Student, (cross) and (single) respectively means cross-split labeling and single-split labeling strategies as introduced in \textsection\ref{sec:reevaluationsetup}.
``Our Best (few-shot)'' is the results achieved by a combination method as introduced in \textsection\ref{sec:bestfewshotperfcombinationo}.
\textbf{Globally best results} for each task are in bold. 
\underline{Best results for minimal few-shot methods} are underlined.
\uwave{Best results for semi-supervised few-shot methods} are marked with wavelines.
}}
\label{tab:debertamainresults}
\end{table*}

\begin{table*}
\small
\centering
\resizebox{\textwidth}{!}{%
  \begin{tabular}{l|cccccccc}
    \toprule[1pt]  
    
     & BoolQ & RTE & WiC & CB & MultiRC & WSC & COPA \\
     \midrule
     Minimal Few-Shot Methods & PET & ADAPET & PET & FlipDA & ADAPET & ADAPET & PET\\
     Training Paradigm & iPET(cross) & Noisy(cross) & iPET(cross) & single & Noisy(cross) &Noisy(single) & iPET(cross)\\
     + MLM & \checkmark & - & - & -& - & - & - \\
    \bottomrule[1pt]
\end{tabular}
}
\caption{\small{The combination of methods that achieves the best few-shot performance for each task. 
There are five minimal few-shot methods and five training paradigms as combined options, as \textsection\ref{sec:bestfewshotperfcombinationo} illustrates.
``+MLM'' means adding an additional MLM loss. 
}}
\label{tab:few-shot combination}
\end{table*}

\begin{table}
    \centering
    \small
    \resizebox{\linewidth}{!}{%
    \begin{tabular}{l|cc|cc|cc}
    \toprule[1pt]
    \multirow{2}*{Methods} 
    & \multicolumn{2}{c|}{RTE} & \multicolumn{2}{c|}{WiC} & 
    \multicolumn{2}{c}{COPA} \\
    & Prev. &   Ours  &  Prev. & Ours &  Prev. & Ours \\
    \midrule
    PET 
    & 69.80 
    & 72.83  
    & 52.40 
    & 53.87 
    & 95.00 
    & 81.75 \\
    \\[-1.5ex]
    \multirow{1}*{ADAPET} 
    & 76.50 & 74.28 & \uwave{54.40} & \uwave{58.07} & \uline{89.00}  & \uline{81.75} \\
    \\[-1.5ex]
    \multirow{1}*{P-tuning}
    & 76.50 & 63.27 & \uwave{56.30} & \uwave{55.49} & \uline{87.00} & \uline{85.25} \\

    \\[-1.5ex]

    \multirow{1}*{FlipDA}
    & 70.67 &70.85 & 54.08 & 57.17 & 89.17 & 87.50 \\
    \\[-1.5ex]
    \multirow{1}*{+MLM}
    & 62.20 & 71.48 & 51.30 & 52.39 & 86.70 & 85.75 \\
    \\[-1.5ex]
    \multirow{1}*{iPET}
    & 74.00 & 72.35 & 52.20 & 54.78 & 95.00 & 84.00 \\
    \bottomrule[1pt]
    \end{tabular}}
    \caption{{
    Comparison of prior evaluations and our evaluation.
    We report the absolute performance of different methods respectively from previous evaluation (Prev.) and our evaluation framework (Ours.) on RTE, WiC and COPA tasks.
    The results are based on ALBERT.
    Results of previous evaluation are taken from the original papers, including ADAPET~\cite{adapet}, P-tuning~\cite{gptunderstandstoo},
    FlipDA~\cite{zhou2021flipda} and iPET~\cite{Schick2021ItsNJ}.
    Since \cite{SchickS21_PET} reported results of PET+MLM on different tasks, we re-experimented on the same tasks under the same setting as~\cite{SchickS21_PET}. {\uwave{Wave lines} and \uline{underlines}} indicate examples of inaccurate estimates of relative gaps in prior works (see text for details).
    }}
    \label{tab:relativeperformance}
\end{table}

\subsection{Few-Shot Methods} \label{sec:few-shot-methods}

We now proceed to re-evaluate state-of-the-art few-shot methods under our evaluation framework with the \solution strategy.
We consider two types: \textit{minimal few-shot methods}, which only assume access to a small labeled dataset, 
including Classification~\cite[CLS;][]{bertpaper},
PET~\cite{Schick2021ItsNJ}, 
ADAPET~\cite{adapet}, P-tuning~\cite{gptunderstandstoo} and FlipDA~\cite{zhou2021flipda}; 
and \textit{semi-supervised few-shot methods}, which allow accessing an additional unlabeled dataset,
including PET+MLM~\cite{SchickS21_PET}, iPET~\cite{Schick2021ItsNJ} and Noisy Student~\cite{XieLHL20_NoisyStudent}.

\subsection{Experimental Setup}\label{sec:reevaluationsetup}

The same benchmark datasets, metrics, and hyper-parameter space as in \textsection\ref{sec:setup} are used.
We use 32 labeled samples for training.
We consider two labeling strategies to obtain the pseudo-labels on unlabeled samples used by the semi-supervised methods for self-training, including \textit{single-split labeling} and \textit{cross-split labeling}.
In the single-split setting~\cite{Schick2021ItsNJ}, pseudo-labels are generated by the models trained on the same data split. 
In the cross-split setting in our evaluation framework, 
the pseudo-labels are generated by the models trained on multiple different data splits. 
More configuration details are in Appendix~\ref{apdx:part3_details}.

\subsection{Main Results and Analysis}
\label{sec:main_results}

\paragraph{Re-Evaluation Results} 
Table~\ref{tab:debertamainresults} shows our re-evaluation results.
The prompt-based fine-tuning paradigm significantly outperforms the classification fine-tuning on all tasks and on both pretrained models (with an advantage of more than 15 points on average). 
DeBERTa outperforms ALBERT consistently. 
We observe significant differences in performance between different prompt-based minimal few-shot methods with ALBERT (e.g., ADAPET and FlipDA outperform PET respectively by about 4 points and 2 points on average) while differences with DeBERTa are slight (e.g., PET, ADAPET, P-tuning, and FlipDA have a performance gap of only about 1.0 points on average). 
In contrast, semi-supervised few-shot methods (i.e., iPET and Noisy) generally improve 1--2 points on average compared to minimal few-shot methods on both models.

\paragraph{Comparison to Prior Evaluations}
Since we have proved that our evaluation framework is more reliable in estimating method performance as shown in Section \ref{sec:settingmainresults},
we conduct experiments to compare the estimates by our evaluation framework and prior evaluations to study whether model performance was accurately estimated in prior work.

Table~\ref{tab:relativeperformance} lists the absolute performance from prior evaluations and our evaluation.
Results show the absolute performance of few-shot methods in prior evaluations was generally overestimated on RTE and COPA.
Similar findings have been highlighted in prior works~\cite{truefew-shot,revisitingfewsamplefinetune}, and our evaluation framework confirms the findings under a more reliable setup. This results from a more reliable evaluation procedure that emphasizes dev-test correlation to prevent overfitting (discussed in Section \ref{sec:howtosplit}).

Besides, the relative gaps between different methods were not accurately estimated by the prior reported numbers.
For example, according to the reported results in prior works, ADAPET outperforms P-Tuning on COPA and P-Tuning beats ADAPET on WiC, while our evaluation reveals the opposite.
On one hand, this is because prior results were obtained under a less reliable evaluation procedure (discussed in Section \ref{sec:howtosplit}). Deviation in the estimates of absolute performance contributes to inaccuracy in the estimates of relative performance. On the other, prior experiments were not conducted under a shared evaluation procedure. These two factors are corrected by our re-evaluation under the more reliable proposed framework.

To sum up, our re-evaluation compares all methods on a common ground, revealing the following:
\begin{theorem}\label{thm:decrease}
The absolute performance and the relative gap of few-shot methods were in general not accurately estimated in prior literature.
This is corrected by our new evaluation framework with improved reliability. 
It highlights the importance of evaluation for obtaining reliable conclusions.
Moreover, the benefits of some few-shot methods (e.g., ADAPET) decrease on larger pretrained models like DeBERTa.
\end{theorem}


\subsection{What is the Best Performance Few-Shot Learning can Achieve?}\label{sec:bestfewshotperfcombinationo}

We further explore the best few-shot performance by combining various methods, and evaluating under our evaluation framework.
For combined options, we consider five minimal few-shot methods (i.e., CLS, PET, ADAPET, P-tuning, and FlipDA), five training paradigms (i.e., single-run, iPET (single/cross), and Noisy Student (single/cross)), and the addition of a regularized loss (+MLM).
We experiment with all possible combinations and report the best for each task.

``Best (few-shot)'' in Table~\ref{tab:debertamainresults} achieves the best results on all tasks among all methods.
Existing few-shot methods can be practically used in combination.
Compared to RoBERTa (fully-sup)~\cite{liu2019roberta}, the performance gap has been further narrowed to 2.89 points on average.\footnote{Note that the gap could be larger since RoBERTa-Large has a smaller number of parameters than DeBERTa, and RoBERTa (fully-sup) does not incorporate additional beneficial techniques such as ensembling or self-training.} Compared to DeBERTa (fully-sup), there is still a sizeable gap between few-shot and fully-supervised systems.

We list the best-performing combination for each task in Table~\ref{tab:few-shot combination}.
The best combinations are very different across tasks, and there is no single method that dominates most tasks. 
PET and ADAPET as well as iPET and Noisy Student are about equally preferred while cross-split labeling and no regularization term perform better. 
We thus recommend future work to focus on the development of methods that achieve consistent and robust performance across tasks.
We summarize the following findings:
\begin{theorem}\label{thm:combinationsota}
Gains of different methods are largely complementary. A combination of methods largely outperforms individual methods, performing close to a strong fully-supervised baseline on RoBERTa. However, there is still a sizeable gap between the best few-shot and the fully-supervised system.
\end{theorem}
\begin{theorem}\label{thm:combination}
No single few-shot method dominates most NLU tasks. This highlights the need for the development of few-shot methods with more consistent and robust performance across tasks.
\end{theorem}

\section{FewNLU Toolkit}

\begin{figure}[ht]
  \centering {\includegraphics[width=\linewidth] {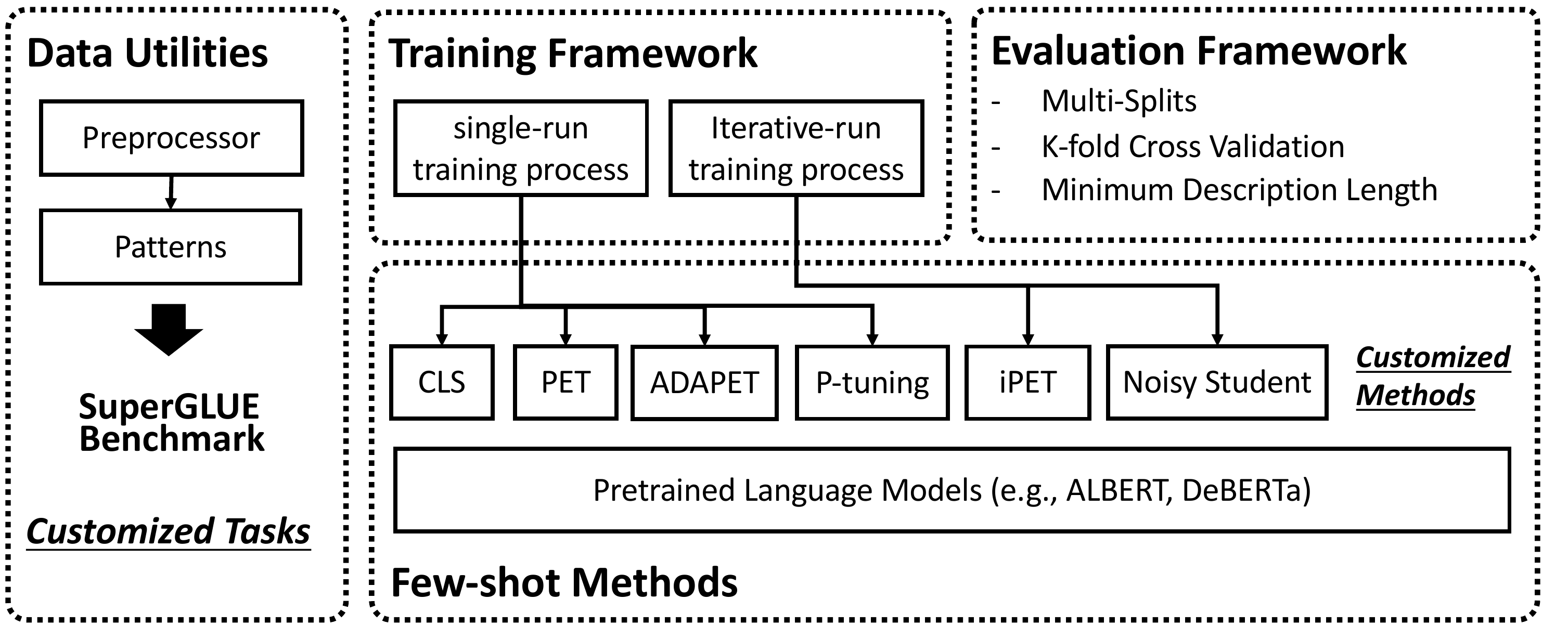}}\\
\caption{{Architecture of FewNLU.}}
\label{fig:toolkit}
\end{figure}

We open-source FewNLU, an integrated toolkit designed for few-shot NLU.
It contains implementations of state-of-the-art methods, data processing utilities, a standardized few-shot training framework, and most importantly, our proposed evaluation framework.
Figure~\ref{fig:toolkit} shows the architecture.
We hope FewNLU could facilitate benchmarking few-shot learning methods for NLU tasks and expendit the research in this field.
\section{Conclusions}

We introduce an evaluation framework, re-evaluate a number of few-shot learning methods under the evaluation framework with a novel \solution strategy, and release a few-shot toolkit.
Apart from this, we also aim at advancing the development of few-shot learning by sharing several new experimental findings.
We identify several new directions for future work:
(1) 
In practice, how to define the hyper-parameter search space a priori is a challenge. 
(2) It is critical for the community to iterate and converge on a common evaluation framework.
(3) Few-shot natural language generation might also be studied in a similar framework.

\section*{Acknowledgements}
We thank Dani Yogatama for valuable feedback on a draft of this paper.
Tang is funded by NSFC for Distinguished Young Scholar (61825602).
Zheng, Ding, Tang, and Yang are funded by the National Key R\&D Program of China (2020AAA0105200) and supported by Beijing Academy of Artificial Intelligence (BAAI).
Zheng is Funded by China Postdoctoral Science Foundation (2021M690471).
Zhou and Li are supported in part by the National Natural Science Foundation of China Grant 62161146004, Turing AI Institute of Nanjing and Xi'an Institute for Interdisciplinary Information Core Technology.





\bibliography{ref}
\bibliographystyle{acl_natbib}

\appendix
\newpage

\section{Appendix}

\subsection{Fixed Hyper-Parameters are not Optimal}\label{sec:doweneedtoselect}

Some prior works~\cite{SchickS21_PET,Schick2021ItsNJ,adapet} perform few-shot learning with a fixed set of hyper-parameters (determined by practical considerations and experiences) without early stopping and any model selection.

\begin{table}[ht]
\centering
\small
\resizebox{\linewidth}{!}{%
\begin{tabular}{c|c|c|c|c|c|c}
\toprule[1pt]
    & \multicolumn{4}{c|}{Hyper-Parameters}  & \multirow{2}*{Test Acc.} & \multirow{2}*{Avg.} \\
    \cline{2-5}
    & P & LR & Step & WR &  \\
    \midrule
        \multirow{5}*{Fixed} 
        & 0 & \multirow{5}*{1e-5} & \multirow{5}*{250} & \multirow{5}*{0} & 69.31 \tiny{$\pm$4.39} &  \multirow{5}*{67.36}\\
        & 1 &&&& 61.13 \tiny{$\pm$0.91} & \\
        & 2 &&&& 63.06 \tiny{$\pm$1.50} & \\
        & 3 &&&& 63.06 \tiny{$\pm$1.82} & \\
        & 4 &&&& 80.26 \tiny{$\pm$1.85} & \\
    \midrule[1pt]
        \multirow{5}*{Optimal}
        & 0 & 1e-5 & 300 & 0.05  & 72.44 \tiny{$\pm$1.85} & \multirow{5}*{70.42}\\
        & 1 & 5e-6 & 300 & 0.05 & 63.78 \tiny{$\pm$1.37} & \\
        & 2 & 5e-6 & 300 & 0 & 69.07 \tiny{$\pm$5.55} & \\
        & 3 & 5e-6 & 300 & 0 & 65.70 \tiny{$\pm$1.25} & \\
        & 4 & 5e-6 & 300 & 0 & 81.11 \tiny{$\pm$1.37} & \\
    \bottomrule[1pt]
\end{tabular}
}
\caption{
\small{
Performance of PET on RTE task with different hyper-parameters. 
The patterns and fixed hyper-parameters are reported by~\cite{Schick2021ItsNJ}.
Base model: DeBERTa-xxLarge,
``P'': pattern ID,
``LR'': learning rate,
``Step'': number of training steps, 
``WR'': warmup ratio.
}}
\label{tab:preliminaryresults}
\end{table}

We first study how well fixed hyper-parameters transfer to a new scenario, e.g. switching to another base pretrained model.
We perform preliminary experiments on FewGLUE with 64 labeled sample
based on DeBERTa.
Firstly, we experiment with the fixed hyper-parameters used for ALBERT in~\cite{Schick2021ItsNJ}.
Secondly, we manually try other hyper-parameters to find out whether there are better configurations.
From Table~\ref{tab:preliminaryresults}, we observe:
\begin{enumerate}
    \item
    Certain factors, especially the patterns, impact the performance a lot (best 80.26\%, and worst 61.13\%). However, we cannot differentiate between them without a development set. 

    \item There exists a hyper-parameter (``Optimal'' in Table~\ref{tab:preliminaryresults}) that performs much better than the fixed one.
    A mechanism to identify the best hyper-parameter setting is thus necessary.

    \item Results show a good hyper-parameter on ALBERT does not work well on DeBERTa.
    Fixed hyper-parameters are not optimal and we need to re-select them given new conditions.
\end{enumerate}

\subsection{Details of How to Construct Data Splits}\label{apdx:part1_details}

\subsubsection{Datasets}

To justify the proposed evaluation framework, we perform experiments on the few-shot SuperGLUE benchmark, which was constructed to include some of the most difficult language understanding tasks for current NLP approaches \cite{Wang2019SuperGLUEAS}.
Unlike other NLU benchmarks (e.g., GLUE~\cite{WangSMHLB19}) that contain single-sentence tasks, SuperGLUE consists of complicated ones that are sentence-pair or sentence-triple tasks, which demand advanced understanding capabilities.
Seven SuperGLUE tasks are considered, including question answering (BoolQ~\cite{BoolQ2019} \& MultiRC~\cite{MultiRC2018}), textual entailment (CB~\cite{de2019commitmentbank} \& RTE~\cite{RTE2005}), word sense disambiguation (WiC~\cite{wic-paper}), causal reasoning (COPA~\cite{COPA2011}), and co-reference resolution (WSC~\cite{WSC2012}).



    

\subsubsection{Hyper-parameters}\label{sec:hyperparameterspace}

To quantitatively evaluate different data-split strategies, we perform extensive experiments with the following hyper-parameter search space.
Data-split experiments are based on DeBERTa-xxLarge.
The hyper-parameter search space is shown in Table \ref{tab:data_split_search_space}.
We use the same prompt patterns as in~\cite{Schick2021ItsNJ}.
To observe the changes of performance and correlation metrics w.r.t different $K$ values, 
we also experimented with $K=\{2, 4, 8, 16\}$ over three tasks (i.e., BoolQ, RTE and COPA).


\begin{table}[htbp]
\small
\centering
\resizebox{\linewidth}{!}{%
  \begin{tabular}{l|ccccccc}
    \toprule[1pt]
    Hyper-parameter & Value \\ 
    \midrule
    Learning Rate & $\{5e-6, 1e-5\}$\\
    Maximum Training Step &  $\{250, 500\}$\\
    Evaluation Frequency & $\{0.02, 0.04\}$\\
    Number of Runs $K$ & 4 \\
    Split Ratio $r$ for \solution & 1:1 \\
    \bottomrule[1pt]
\end{tabular}
}
\caption{Hyper-parameter Search Space for Data-Split Strategy Evaluation}
\label{tab:data_split_search_space}
\end{table}


\subsubsection{Evaluation Results with 32 Labeled Data}
\label{apdx:strategy_32}

In the data-split strategy evaluation, in addition to the 64-data-setting results in the main text, we also experimented with 32 labeled data as~\cite{Schick2021ItsNJ,SchickS21_PET,adapet}.
The 32-data-setting results are also provided in Table~\ref{tab:fewshotperformance-32}.

\subsubsection{Leave-One-Out Cross Validation Results}\label{apdx:loocv}

We also experiment with another useful data split strategy, leave-one-out cross validation (LOOCV).
In fact, LOOCV is a special case of $K$-fold cross validation when $K$ equals the number of labeled data.
Since LOOCV takes even longer time than any other data split strategies, we only experimented on three tasks, including BoolQ, RTE and WiC tasks.
Both performance and correlation results are shown in Table~\ref{tab:loocv}.
Our results show that compared to other strategies, LOOCV achieved worse test performance as well as correlation. 
LOOCV only uses a single instance for validation each time, and thus leads to poor correlation and random model selection. As a result, the performance estimation is subject to much randomness.
\begin{table}[]
    \centering
    \small
    \resizebox{0.9\linewidth}{!}{%
    \begin{tabular}{l|l|ccc}
        \toprule[1pt]
        & &  BoolQ & RTE & WiC \\
        \midrule
        \multirow{3}*{Multi-Splits}
        & Perf. 
        & \makecell{82.67 \\ \std0.78}
        & \makecell{79.42 \\ \std2.41}
        & \makecell{67.20 \\ \std1.34}
        \\ 
        \cline{2-5}\\[-1.5ex]
        & Corr. 
        & 0.7079 & 0.8266 & 0.9464 
        \\
        \midrule
        \multirow{3}*{CV}
        & Perf. 
        & \makecell{82.71 \\ \std1.29}
        & \makecell{77.80 \\ \std2.25}
        & \makecell{64.42 \\ \std1.63}
        \\
        \cline{2-5}\\[-1.5ex]
        & Corr. & -0.0497 & 0.8561 & 0.8184\\
        \midrule
        \multirow{3}*{LOOCV} 
        & Perf. 
        &  \makecell{80.20 \\ \std5.63} 
        & \makecell{63.91\\\std5.37} 
        & \makecell{62.40 \\ \std4.70} \\
        \cline{2-5}\\[-1.5ex]
        & Corr. & -0.8001 & -0.5070 & 0.1998 \\
        \bottomrule[1pt]
    \end{tabular}}
    \caption{Test performance and correlation results of leave-one-out cross validation on BoolQ, RTE and WiC tasks with 64 labeled examples.}
    \label{tab:loocv}
\end{table}

\begin{table*}[t] 
\small
\centering
\begin{subtable}{\textwidth}
\centering
\caption{Results of test performance of the selected hyper-parameter.}
\label{tab:fewshotperformance:tab64}
\resizebox{0.8\textwidth}{!}{%
  \begin{tabular}{l|cccccccccc}
    \toprule[1pt]  
    \multirow{2}*{}
    & BoolQ & RTE & WiC & \multicolumn{2}{c}{CB} & \multicolumn{2}{c}{MultiRC} & WSC & COPA & Avg. \\
    & Acc. & Acc. & Acc. & Acc. & F1 & F1a & EM. &  Acc. & Acc \\
    \midrule  
    \multirow{2}*{CV} & 77.29 & 75.63 & 55.56 & \textbf{89.29} & \textbf{80.66} & \textbf{78.61} & \textbf{42.26} & \textbf{78.37} & \textbf{90.00} & \multirow{2}*{\textbf{74.61}} \\
    & \std 3.32 &\std 4.26 &\std {1.06} &\std \textbf{3.86} &\std 14.87 &\std \textbf{0.84} &\std 2.07 &\std 4.26 &\std 2.45 & \\
    \\[-1.5ex]
    \multirow{2}*{MDL} & \textbf{79.29} & 75.87 & 53.53 & 79.61 & 59.25 & 75.77 & 37.30 & 77.82 & 76.25 & \multirow{2}*{69.82} \\
    &  \std 6.01 &\std5.19 &\std\textbf{0.58} &\std5.42 &\std11.27 &\std4.72 &\std6.27 &\std4.19 &\std12.50 & \\
    \\[-1.5ex]
    \multirow{2}*{\solution} 
    & 78.11 & \textbf{79.42} & \textbf{61.72} & 83.04 & 70.93 & 78.23 & 41.45 & 74.52 & 84.75 & \multirow{2}*{73.62} \\
    & \std 2.63	&\std \textbf{1.79} &\std 3.10 &\std 6.66 &\std 13.40 &\std 1.24 &\std \textbf{1.74} &\std 3.96 &\std \textbf{2.12} & \\
    \bottomrule[1pt]
\end{tabular}
}
\end{subtable}
\begin{subtable}{\textwidth}
\centering
\caption{Results of correlation between the development and training sets.}
\label{tab:correlationresults:tab32}
\resizebox{0.8\textwidth}{!}{%
  \begin{tabular}{l|cccccccc}
    \toprule[1pt]  
     & BoolQ & RTE & WiC & CB & MultiRC & WSC & COPA & Avg. \\
    \midrule  
    CV & 0.4134 & 0.6759 & 0.4189 & 0.0938 & 	0.1061 & -0.1683 & {0.6567} & 0.3138 \\
    \\[-1.5ex]
    MDL & \textbf{0.6394} & 0.5687 & -0.0732 & 0.2127 & 	0.1690 & 0.0741 & 0.1100 & 0.2429 \\
    \\[-1.5ex]
    \solution 
    & 0.5347 & {0.6911} & \textbf{0.8448} & \textbf{0.7232} & \textbf{0.6280} & \textbf{0.0853} & 0.4531 & \textbf{0.5657}\\
    \bottomrule[1pt]
\end{tabular}
}
\end{subtable}\\
\caption{\small{Evaluation results of different few-shot data-split strategies with PET on FewGLUE ($K$=4) under the same data setting as~\cite{Schick2021ItsNJ,SchickS21_PET,adapet} with 32 labeled data.
Larger scores indicate that a data-split strategy effectively selects a model that achieves better test-set performance.
The best results for each task are denoted in bold.}}
\label{tab:fewshotperformance-32}
\end{table*}


\subsection{How to Define the Hyper-parameter Search Space}\label{sec:howtosearchparam}

Aside from how to construct the data splits, another important question for the evaluation framework is how to define the hyper-parameter search space. We left this question in the future work. However, we did several preliminary experiments that could reveal certain insights into the problem.

\subsubsection{Should We Search Random Seeds?}\label{sec:keyfactor}

We focus on two types of factors that affect few-shot evaluation, hyper-parameters and randomness.
Randomness could cause different weight initialization, data splits, and data order during training.
Empirically, how randomness is dealt with differs depending on the use case. In order to obtain the best possible performance, one could search over sensitive random factors such as random seeds.
However, as we focus on benchmarking few-shot NLU methods, we report mean results (along with the standard deviation) in our experiments in order to rule out the effects of randomness and reflect the average performance of a method for fair comparison and measurement.

\subsubsection{Experiments}\label{sec:crucialhyperparam}

\paragraph{Experimental Setup}

To examine how a certain factor affects few-shot performance,
we assign multiple different values to a target factor while fixing other hyper-parameters. 
We report the standard deviation over the multiple results.
Larger values indicate that a perturbation of the target factor would largely influence the few-shot performance and the factor thus is crucial for searching.
We experiment on BoolQ, RTE, CB, and COPA tasks.
Considered factors include: sample order during training, prompt pattern, training batch size, learning rate, evaluation frequency, and maximum train steps.
\begin{table}[!h]
\small
\centering
\resizebox{\linewidth}{!}{%
  \begin{tabular}{c|l|cccc}
    \toprule[1pt]  
    & {Hyper-params} &BoolQ & RTE & COPA & {CB} \\
    \midrule
    \multirow{5}*{\shortstack{Dev\\Set}}
    
    & Train Order 
    & \textbf{3.64}  
    & \textbf{4.01}  
    & \textbf{2.17}  
    & \textbf{2.21/6.09} \\
    
    & Prompt Pattern 
    & \textbf{3.44}  
    & \textbf{10.28}	 
    & \textbf{5.80} 
    & \textbf{3.18/4.07}	\\
    
    & Train Batch
    & \textbf{3.34} 
    &	1.33
    &	\textbf{2.64}  
    & 1.01/\textbf{5.87} \\
    
    & Learning Rate 
    & 0.00  
    & 1.63 
    & 1.97  
    & 1.56/\textbf{4.56}\\
    
    &Eval Freq 
    &\textbf{ 2.39}  
    & \textbf{2.96}  
    & \textbf{2.73} 
    & 0.45/0.82 \\
    \midrule
    \multirow{5}*{\shortstack{Test \\ Set}}
    & Train Order 
    & 0.87 
    & 1.87
    & \textbf{2.17}  
    & \textbf{3.01/4.73} \\

    & Prompt Pattern 
    & \textbf{2.85}  
    & \textbf{10.03} 
    & \textbf{2.65} 
    & \textbf{6.45/7.08} \\
    
    & Train Batch 
    & \textbf{2.44}
    & 1.09	
    & 0.72
    & 0.89/1.32 \\
    
    & Learning Rate 
    & 0.17  
    &	0.65  
    &	0.52 
    &	\textbf{4.82/7.25}  \\
    
    & Eval Freq 
    & 0.84  
    & 0.53  
    & 1.18  
    & 0.77/\textbf{2.07}  \\

    \bottomrule[1pt]
\end{tabular}
}
\caption{\small{
Analysis of different factors on BoolQ, RTE, CB and COPA using PET and DeBERTa. 
The metric is standard deviation.
Hyper-parameters are set the best-performing ones obtained in \textsection\ref{sec:reevaluation} while the target factor is assigned with multiple values.
``Train Order'': training sample order;
``Train Batch'': total train batch size;
``Eval Freq'': evaluation frequency.}}
\label{tab:factors}
\end{table}

\paragraph{Results and Analysis}
Results are in Table~\ref{tab:factors}. We mark values larger than a threshold of $2.0$ in bold.
We can see that the prompt pattern is the most influential factor among all, indicating the design or selection of prompt patterns is crucial.
Training example order also significantly affects the performance.
The evaluation frequency affects the score on the small development but not on the test set. 
We speculate that a lower frequency selects a model with better performance on the small development set, but the gains do not transfer to the test set because of partial overfitting. To conclude:
\begin{theorem}\label{thm:crucialfactor}
We recommend to at least search over prompt patterns during hyper-parameter tuning, and it is also beneficial to search others.
All comparison methods should be searched and compared under the same set of hyper-parameters.
\end{theorem}

\begin{table}[htbp]
\small
\centering
\resizebox{\linewidth}{!}{%
  \begin{tabular}{l|ccccccc}
    \toprule[1pt]
    Hyper-parameter & Value \\ 
    \midrule
    Learning Rate & $\{6e-6, 8e-6, 1e-5\}$\\
    Evaluation Frequency & $\{0.02, 0.04, 0.08\}$\\
    Training Batch Size & $\{8, 16, 32, 64\}$\\
    Sample Order Seed & $\{10, 20, 30, 40, 50, 60, 70, 80\}$\\
    \bottomrule[1pt]
\end{tabular}}
\caption{Hyper-parameter Search Space for Crucial Factor Evaluation}
\label{tab:crucial_factor_space}
\end{table}

\subsubsection{Detailed Configuration}
\label{apdx:part2_details}


For a given task and a target factor, we fixed the hyper-parameters to be the best-performing ones obtained in Section~\ref{sec:howtosplit}, and assigned multiple values for the target factor.
For the prompt pattern, we assigned it with the same values as~\cite{Schick2021ItsNJ}.
Possible values for other hyper-parameters are in Table \ref{tab:crucial_factor_space}.

\begin{table*}
\small
\centering
\resizebox{0.7\textwidth}{!}{%
  \begin{tabular}{l|cccccccc}
    \toprule[1pt]  
     & BoolQ & RTE & WiC & CB & MultiRC & WSC & COPA \\
     \midrule
     Learning Rate & 1e-5 & 5e-6 & 5e-6 & 1e-5 & 1e-5 & 1e-5 & 1e-5\\
     Maximum Training Step & 250 & 250 & 250 & 250 & 250 & 500 & 500 \\
     Evaluation Frequency & 0.02 & 0.02 & 0.02 & 0.02 & 0.04 & 0.04 & 0.02 \\
     Prompt Pattern & 1 & 5 & 2 & 5 & 1 & 2 & 0 \\
    \bottomrule[1pt]
\end{tabular}}
\caption{The best hyper-parameters searched for PET. We search each task with a learning rate of \{1e-5,5e-6\},  max steps of \{250,500\}, evaluation frequency ratio of \{0.02,0.04\}, and all the available prompt patterns. Therefore, each task has $8N$ hyper-parameter combinations, where $N$ is the number of available prompt patterns, i.e., 6 for BoolQ and RTE, 3 for WiC, and 2 for COPA. }
\label{tab:pet_hyperparameter}
\end{table*}

\begin{table*}
\small
\centering
  \begin{tabular}{l|ccccccc}
    \toprule[1pt]  
     & BoolQ & RTE & WiC & CB & MultiRC & WSC & COPA \\
     \midrule
     Learning Rate & 1e-5 & 5e-6 & 5e-6 &1e-5 & 5e-6 & 5e-6 & 5e-6 \\
     Maximum Training Step & 250 & 500 & 500 & 500 & 500 & 250 & 500\\
     Evaluation Frequency & 0.04 & 0.04 & 0.02 & 0.02 & 0.02 & 0.04 & 0.04 \\
     Prompt Pattern & 1 & 5 & 2 & 5 & 0 & 1 & 0\\
    \bottomrule[1pt]
\end{tabular}
\caption{The best hyper-parameters searched for ADAPET. We search each task with a learning rate of \{1e-5,5e-6\},  max steps of \{250,500\}, evaluation frequency ratio of \{0.02,0.04\}, and all the available prompt patterns. Therefore, each task has $8N$ hyper-parameter combinations, where $N$ is the number of available prompt patterns, i.e., 6 for BoolQ and RTE, 3 for WiC, and 2 for COPA. }
\label{tab:adapet_hyperparameter}
\end{table*}

\begin{table*}[htbp]
\small
\centering
  \begin{tabular}{l|cccccccc}
    \toprule[1pt]  
     & BoolQ & RTE & WiC & CB & MultiRC & WSC & COPA \\
     \midrule
     Learning Rate & 5e-6 & 5e-6 & 5e-6 & 1e-5 & 1e-5 & 5e-6 & 1e-5\\
     Maximum Training Step & 500 & 250 & 500 & 250 & 500 & 500 & 500 \\
     Warmup Ratio & 0.0 & 0.0 & 0.0 & 0.1 & 0.1 & 0.1 & 0.1 \\
     Evaluation Frequency & 0.02 & 0.02 & 0.02 & 0.04 & 0.02 & 0.02 & 0.04\\
     Prompt Encoder Type & mlp & lstm & lstm & lstm & lstm & lstm & mlp\\
    \bottomrule[1pt]
\end{tabular}
\caption{The best hyper-parameters searched for P-tuning.We search each task with a learning rate of \{1e-5,5e-6\},  max steps of \{250,500\}, warmup ratio of \{0.0,0.1\}, evaluation frequency ratio of \{0.02,0.04\}, and prompt encoder implemented with \{``mlp'', ``lstm''\}. }
\label{tab:ptuning_hyperparameter}
\end{table*}

\begin{table*}
\small
\centering
\centering
  \begin{tabular}{l|ccccccc}
    \toprule[1pt]  
     & BoolQ & RTE & WiC & CB & MultiRC & WSC & COPA \\
     \midrule
     Learning Rate & 5e-6 & 1e-5 & 5e-6 &1e-5 & 1e-5 & 5e-6 & 1e-5 \\
     Maximum Training Step & 250 & 500 & 250 & 250 & 500 & 250 & 500\\
     Evaluation Frequency & 0.04 & 0.04 & 0.04 & 0.04 & 0.02 & 0.02 & 0.04 \\
     Prompt Pattern & 0 & 5 & 2 & 5 & 0 & 0 & 0 \\
     Generation Method & sample & greedy & sample & greedy & greedy & sample & greedy\\
     Drop Inconsistant Data & - & \checkmark & - & - &  \checkmark& - & \checkmark \\
    \bottomrule[1pt]
\end{tabular}
\caption{The best hyper-parameters searched for FlipDA. We search three generation methods, try dropping inconsistant data or not. We search each task with a learning rate of \{1e-5,5e-6\},  max steps of \{250,500\}, evaluation frequency ratio of \{0.02,0.04\}, and all the available prompt patterns. Therefore, each task has $8N$ hyper-parameter combinations, where $N$ is the number of available prompt patterns, i.e., 6 for BoolQ and RTE, 3 for WiC, and 2 for COPA. }
\label{tab:flipda_hyperparameter}
\end{table*}

\subsection{Details of Re-Evaluation}\label{apdx:part3_details}

\subsubsection{Methods}
The five considered minimal few-shot methods are introduced as follows.
\begin{enumerate}
    \item \textbf{Classification} is a conventional finetuning algorithm, which uses the hidden states of a special [CLS] token for classification. 
    \item \textbf{PET} is a prompt-based finetuning algorithm.
    It transforms NLU problems into cloze problems with prompts, and then converts the cloze outputs into the predicted class.

    \item \textbf{ADAPET} is based on PET and decouples the losses for the label tokens. 
    It proposes a label-conditioned masked language modeling (MLM) objective as a regularization term. 
    \item \textbf{P-tuning} is also based on PET and automatically learns continuous vectors as prompts via gradient update.
    \item \textbf{FlipDA} is similar to PET but uses both labeled data and augmented data for training. The augmented data are automatically generated by taking labeled data as inputs. \footnote{In our experiments, we use the best checkpoints searched with PET as the classifier for data selection.}
    
\end{enumerate}
The three semi-supervised few-shot methods are introduced as follows.
\begin{enumerate}
    \item \textbf{PET+MLM} is based on PET and additionally adds an auxiliary language modeling task performed on unlabeled dataset.
    It was first proposed by~\cite{SchickS21_PET} to resolve catastrophic forgetting.
    \item \textbf{iPET} is a self-training method. 
    It iteratively performs PET for multiple generations.
    At the end of each generation, unlabeled data are assigned with pseudo-labels by the fully-trained model, and will be used for training along with train data in the next generation.

    \item \textbf{Noisy Student} is similar to iPET with the difference that Noisy Student injects noises into the input embeddings of the model.

\end{enumerate}

\subsubsection{Hyper-parameter Search Space}
\label{apdx:re-evaluation-search-space}

The hyper-parameter search space for other few-shot methods are shown in Table \ref{tab:re-evaluation_space}.

\begin{table}[htbp]
\small
\centering
\setlength{\tabcolsep}{1mm}
\centering
  \begin{tabular}{c|ccccccc}
    \toprule[1pt]
    Method & Hyper-Parameter & Value \\ 
    \midrule
    \multirow{3}*{CLS} 
    & Learning Rate (DeBERTa) & $\{1e-5, 5e-6\}$\\
    & Learning Rate (ALBERT) & $\{1e-5, 2e-5\}$ \\
    & Maximum Training Step & $\{2500, 5000\}$\\
     \midrule
    \multirow{4}*{\shortstack{PET/\\ADAPET}}
    &Learning Rate (DeBERTa)& $\{5e-6, 1e-5\}$\\
    &Learning Rate (ALBERT) & $\{1e-5, 2e-5\}$ \\
    &Maximum Training Step &  $\{250, 500\}$\\
    &Evaluation Frequency & $\{0.02, 0.04\}$\\
    \midrule
    \multirow{6}*{\shortstack{P-tuning}}
    &Learning Rate (DeBERTa)& $\{5e-6, 1e-5\}$\\
    &Learning Rate (ALBERT) & $\{1e-5, 2e-5\}$ \\
    &Maximum Training Step &  $\{250, 500\}$\\
    &Evaluation Frequency & $\{0.02, 0.04\}$\\
    & Warmup Ratio & $\{0.0, 0.1\}$\\
    & Prompt Encoder Type & \{mlp, lstm\}\\
     \midrule
    \multirow{8}*{\shortstack{FlipDA}}
    &Learning Rate (DeBERTa)& $\{5e-6, 1e-5\}$\\
    &Learning Rate (ALBERT) & $\{1e-5, 2e-5\}$ \\
    &Maximum Training Step &  $\{250, 500\}$\\
    &Evaluation Frequency & $\{0.02, 0.04\}$\\
    &DA Method & \{greedy,sample,beam\}\\
    & Drop Inconsistant Data & \{yes, no\}\\
    &Mask Ratio & Fixed \footnotemark\\
    &Fill-in Strategy & Fixed \footnotemark[9] \\
    \midrule
    \multirow{5}*{ \shortstack{iPET/\\Noisy}} & Unlabeled Data Number & 500\\
    & Increasing Factor & $3.0$ \\
    & Sample Ratio (single-split) & $1.0$ \\
     & Sample Ratio (cross-split) & $2$/$3$\\
    & Dropout Rate for Noisy & $0.05$ \\
    \bottomrule[1pt]
\end{tabular}
\caption{Hyper-parameter Space for Re-Evaluation}
\label{tab:re-evaluation_space}
\end{table}
\footnotetext{As recommended in \cite{zhou2021flipda}, we fix one mask ratio for each dataset, i.e., 0.3 for BoolQ, MultiRC, and WSC, 0.5 for RTE and CB, and 0.8 for COPA and WiC. We fix one fill-in strategy for each dataset, i.e., ``default'' for BoolQ, RTE, WiC, CB, and WSC, ``rand\_iter\_10'' for MultiRC, and ``rand\_iter\_1'' for COPA.}

\subsubsection{The Searched Best Hyper-parameters}
\label{apdx:hyper_params_selection}

We list the searched best hyper-parameter configuration for different tasks and methods in Table~\ref{tab:pet_hyperparameter}, Table~\ref{tab:adapet_hyperparameter}, Table~\ref{tab:ptuning_hyperparameter}, Table~\ref{tab:flipda_hyperparameter}.


\subsubsection{More Discussion on ADAPET}

\begin{figure*}[!t]
    \centering{\includegraphics[width=0.85\linewidth] {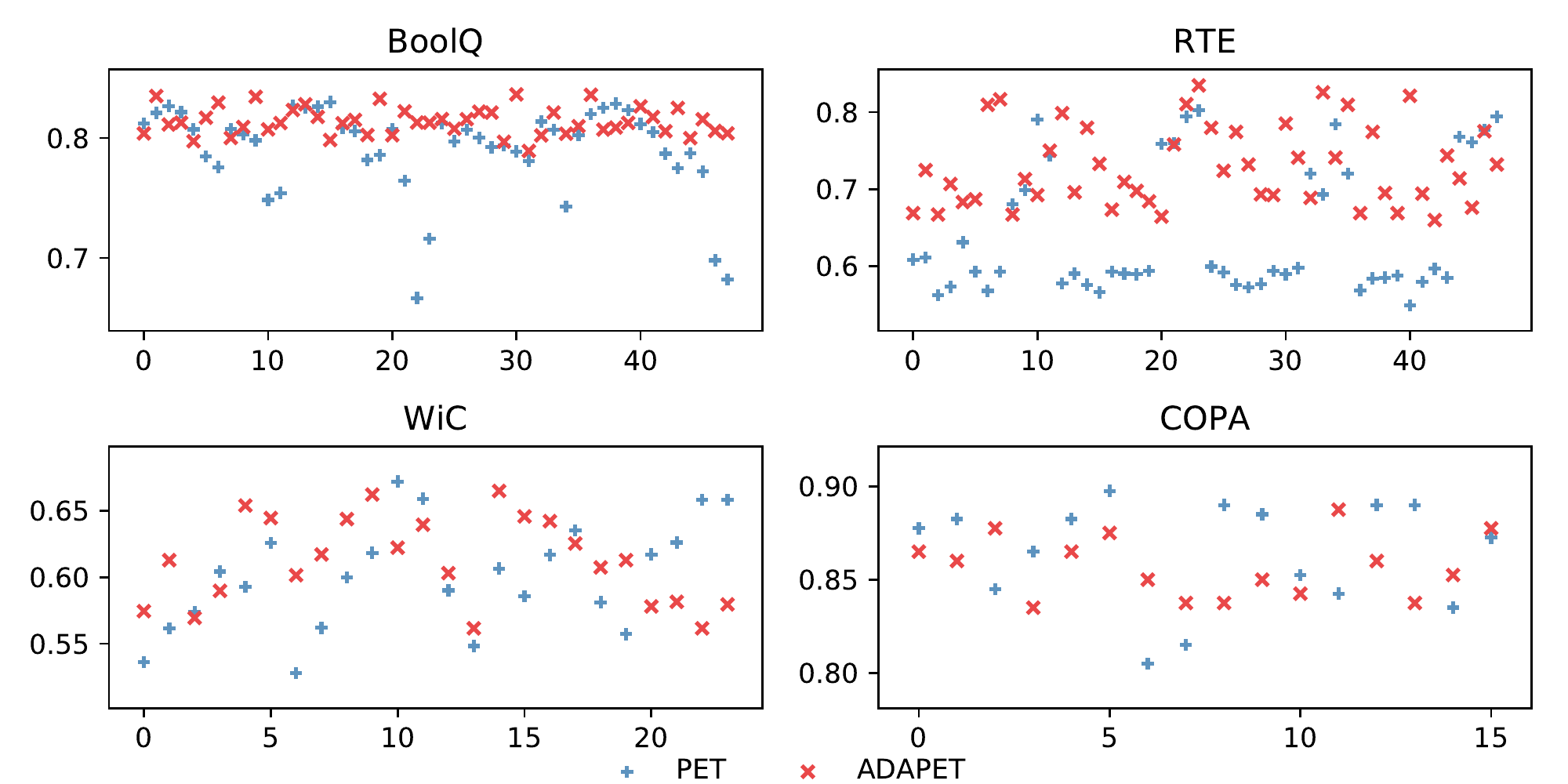}}
    \caption{Visualization of few-shot performance over the same hyper-parameter space of ADAPET and PET based on DeBERTa and \solution. 
    The x-axis is the index of the hyper-parameter combination. 
    We search each task with a learning rate of 1e-5 or 5e-6,  max steps of 250 or 500, evaluation ratio of 0.02 or 0.04, and all the available prompt patterns. Therefore, each task has $8N$ hyper-parameter combinations, where $N$ is the number of available prompt patterns, i.e., 6 for BoolQ and RTE, 3 for WiC, and 2 for COPA. The y-axis is the score of each task given a certain hyper-parameter combination.}
    \label{fig:ADAPET}
\end{figure*}

Since it is observed ADAPET shows less improvement on DeBERTa than it has achieved on ALBERT, we further discuss the phenomena by raising the question what other differences it has made.
We respectively visualize the few-shot performance distribution over the same hyper-parameter space of PET and ADAPET in Figure \ref{fig:ADAPET}.
We observe that PET is more likely to obtain extremely bad results on BoolQ and RTE, while ADAPET shows stable results.
It suggests that ADAPET appears to be more robust to the hyper-parameters, and overall achieves good performance regardless of hyper-parameter selection.
However, ADAPET is less inclined to produce better peak results.
To sum up, we can conclude: Loss regularization (e.g., ADAPET \cite{adapet}) enhances stability w.r.t. hyper-parameters.

\subsubsection{More Discussion on Semi-supervised Few-shot Methods}\label{sec:selftrainingdisucssion}

\begin{table}[!t]
    \small
    \centering
    \setlength{\tabcolsep}{1mm}
    \begin{tabular}{l|c|c|c|c}
    \toprule[1pt]
    task & method & g1 & g2 & g3 \\
    \midrule
    \multirow{2}*{WiC}
    & Multi-Patterns & 60.11 \tiny{$\pm$5.64} & 60.19 \tiny{$\pm$4.12} & 59.66 \tiny{$\pm$4.27} \\
    & Best-Pattern & 64.21 \tiny{$\pm$2.58} & 64.18 \tiny{$\pm$4.61} & 63.37 \tiny{$\pm$6.29} \\

    \midrule
    \multirow{2}*{RTE}
    & Multi-Patterns & 65.08 \tiny{$\pm$10.07} & 69.20 \tiny{$\pm$7.13} & 71.46 \tiny{$\pm$5.59}\\
    & Best-Pattern & 79.39 \tiny{$\pm$2.75} & 81.95 \tiny{$\pm$1.04} & 83.12 \tiny{$\pm$1.42} \\

    \bottomrule
    \end{tabular}
    \caption{\small{The performance results of iPET on both WiC and RTE at every generation (g1, g2, and g3). Each experiment uses either ensemble over all patterns (Multi-Patterns) or ensemble over the only best pattern (Best-Pattern). This experiment is conducted with 1000 unlabeled data and an increasing factor 5.}}
    \label{tab:ensemblepatterns}
\end{table}

We focus on semi-supervised methods that iteratively augment data (i.e., iPET and Noisy Student), which have demonstrated promising results on both models in Table~\ref{tab:debertamainresults}.
Several key points for their success are especially discussed.
\begin{enumerate}
    \item For semi-supervised methods such as iPET and Noisy Student, it is time-consuming when searching over a large hyper-parameter space for each generation.
    We directly use the searched best hyper-parameters for PET in each generation.
    From Table~\ref{tab:debertamainresults}, we can see that their results show advantages over PET (by more than 1 points). 
    It suggests that the best hyper-parameters can be transferred to such methods, to reduce the cost of time and computational resources. 
    If we search for each generation, results might be even better. 
    
    \item Comparing the single-split labeling strategy, the cross-split labeling strategy works better. 
    As the results show, both iPET (cross) and Noisy (cross) outperform iPET (single) and Noisy (single) in most tasks on both models.
    
    \item Another simple and effective technique is our proposed ensemble labeling strategies.
    ~\cite{Schick2021ItsNJ} utilizes the ensemble results over all patterns to label unlabeled data, since it is hard to select patterns.
    Under the \solution strategy, self-training methods can recognize the best pattern, and only ensemble trained models for the best pattern when labeling unlabeled data. 
    Table~\ref{tab:ensemblepatterns} shows the results of iPET on WiC and RTE tasks, respectively ensemble over multiple patterns or ensemble over the only best pattern.
    We can see that results of ensemble with the best pattern significantly outperform results of ensemble with all patterns at every generation.

\end{enumerate}

\end{document}